\documentclass[runningheads]{llncs}
\usepackage{graphicx}
\usepackage{todonotes}
\usepackage{wrapfig}
\usepackage{amssymb}
\usepackage{amsmath}
\usepackage{subfig}

\makeatletter
\renewcommand*{\@fnsymbol}[1]{\ensuremath{\ifcase#1\or \dagger\or \dagger\or \ddagger\or
    \mathsection\or \mathparagraph\or \|\or **\or \dagger\dagger
    \or \ddagger\ddagger \else\@ctrerr\fi}}
\makeatother

\makeatletter
\newcommand{\printfnsymbol}[1]{%
  \textsuperscript{\@fnsymbol{#1}}%
}
\makeatother

\usepackage{ulem}

\begin{document}
%
\title{Dynamic memory to alleviate catastrophic forgetting in continuous learning settings}

\titlerunning{Dynamic memory to alleviate catastrophic forgetting}
%
\author{Johannes Hofmanninger \inst{1} \thanks{Authors contributed equally} 
\and Matthias Perkonigg \inst{1} \printfnsymbol{1} 
\and James A. Brink \inst{2}
\and Oleg Pianykh \inst{2}
\and Christian Herold \inst{1} 
\and Georg Langs \inst{1}}

\authorrunning{J. Hofmanninger, M. Perkonigg et al.}

\institute{Department of Biomedical imaging and Image-guided Therapy, Computational Imaging Research Lab, Medical University of Vienna, Austria
\email{\{johannes.hofmanninger, matthias.perkonigg, georg.langs\}@meduniwien.ac.at}\\
\and
Department of Radiology, Massachusetts General Hospital, Harvard Medical School, Boston, USA
}
\maketitle              
\begin{abstract}
In medical imaging, technical progress or changes in diagnostic procedures lead to a continuous change in image appearance. Scanner manufacturer, reconstruction kernel, dose, other protocol specific settings or administering of contrast agents are examples that influence image content independent of the scanned biology. Such domain and task shifts limit the applicability of machine learning algorithms in the clinical routine by rendering models obsolete over time. Here, we address the problem of data shifts in a continuous learning scenario by adapting a model to unseen variations in the source domain while counteracting catastrophic forgetting effects. Our method uses a dynamic memory to facilitate rehearsal of a diverse training data subset to mitigate forgetting. We evaluated our approach on routine clinical CT data obtained with two different scanner protocols and synthetic classification tasks. Experiments show that dynamic memory counters catastrophic forgetting in a setting with multiple data shifts without the necessity for explicit knowledge about when these shifts occur.

\keywords{Continuous learning  \and Domain adaptation  \and Dynamic memory.}
\end{abstract}
\section{Introduction}
In clinical practice, medical images are produced with continuously changing policies, protocols, scanner hardware or settings resulting in different visual appearance of scans despite the same underlying biology. In most cases, such a shift in visual appearance is intuitive for clinicians and does not lessen their ability to assess a scan. However, the performance of machine-learning based methods can deteriorate significantly when the data distribution changes. \textit{Continuous learning} adapts a model to changing data and new tasks by sequentially updating the model on novel cases. The ground truth labels of these cases can be, for instance, acquired by corrective labelling. 
However, \textit{catastrophic forgetting} is a major undesired phenomenon affecting continuous learning approaches \cite{McCloskey1989CatastrophicProblem}. That is, when a model is continuously updated on a new task or a different data distribution, the model performance will deteriorate on the preceding tasks. Alleviating catastrophic forgetting is one of the major challenges in continuous learning.

\begin{figure}[t!]
    \centering
    \includegraphics[width=1\textwidth]{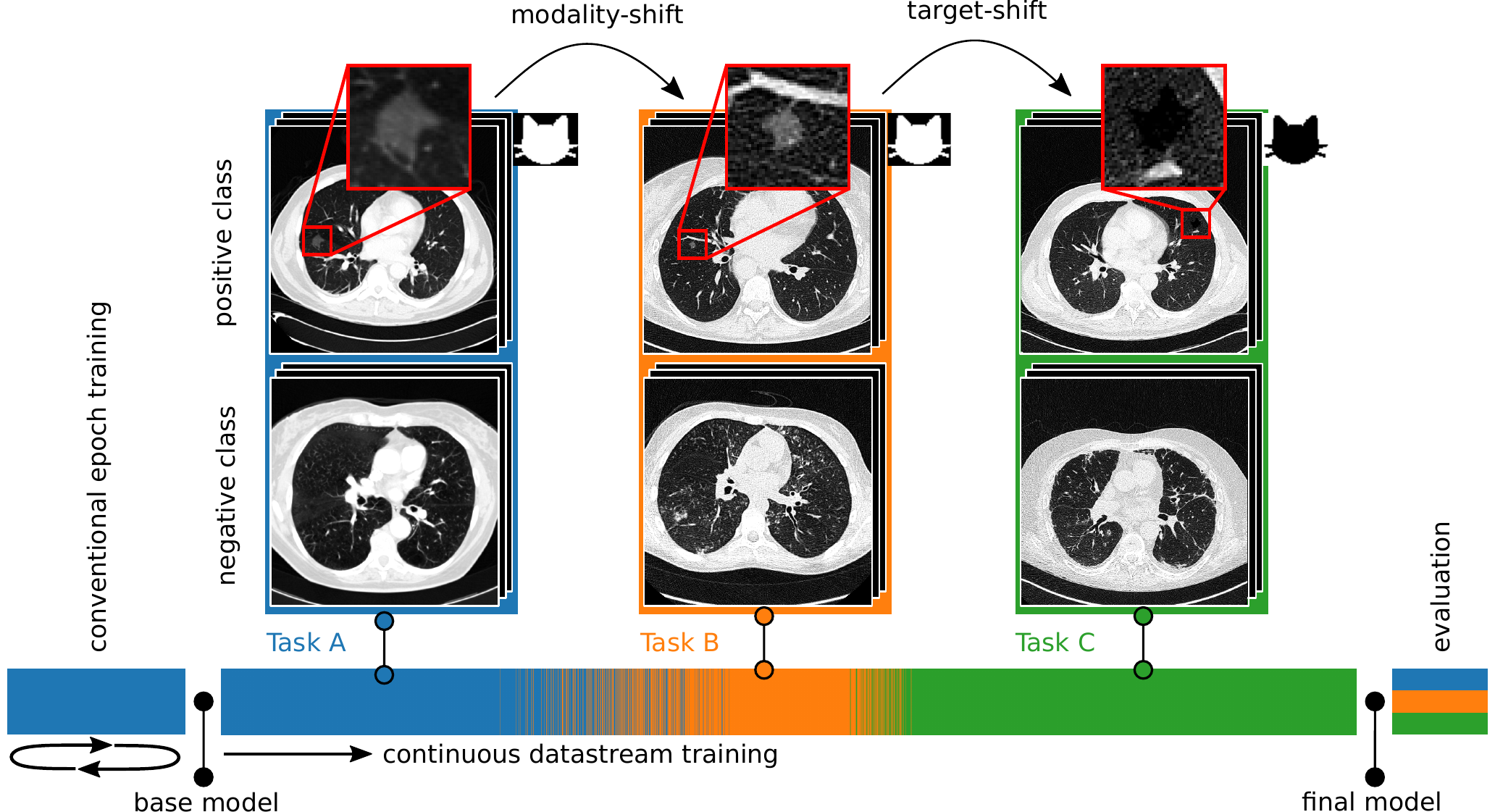}
    \caption{\textbf{Experimental setup for a continuous learning scenario:}
    We assume a learning scenario for which a conventionally trained model (e.g. multi-epoch training)  performing well on Task A is available. This model is continuously updated on a data stream with a shift in overall image appearance caused by scanner parameters (modality-shift) and a shift in the appearance of the classification target (target-shift). The timing of shifts is not known a priori. For evaluation, the final model is evaluated on a test set of all three tasks.}
    \label{fig:exp_setup}
\end{figure}

Here, we propose an approach for a scenario where new data are sequentially available for model training. We aim to utilize such a data stream to frequently update an existing model without the requirement of keeping the entire data available.
We assume a real-world clinical setting where changing visual domains and appearance of classification targets can occur gradually over time and where the information about such an eventual shift in data is not available to the continuous learning system. 
Figure \ref{fig:exp_setup} illustrates the scenario and the experimental setup. A trained model (base-model) has been trained to perform well on a certain classification task. Subsequently, a continuous data stream is used to update the model with the aim to learn variations of the initial task given new data. This model should become accurate on new data, while at the same time staying accurate on data generated by previous technology. Accordingly, the final model is then evaluated on all tasks to assess the effect of catastrophic forgetting and the classification performance. Note that here we use the term \textit{task} to denote the detection of the same target but on shifted visual domains and not for additional target classes.

\paragraph{Related Work}
There are various groups of methods dealing with the problem of continuously updating a machine learning model over time such as continuous learning, continuous domain adaptation or active learning. They operate on similar but different assumptions about the problem settings, the data available and the level of supervision required. For example, continuous domain adaption assumes novel data to be shifted but closely related to the previous data. Continuous learning makes no assumptions about domain shifts and is not limited to a specific task to perform in new data (e.g. incremental learning). Active learning is characterized by the task of automatically selecting examples for which supervision is beneficial. In this work, we propose a continuous learning technique which can also be categorized as a supervised continuous domain adaptation method.

Various methods for continuous learning have been proposed to alleviate catastrophic forgetting in scenarios where multiple tasks are learned sequentially.

A popular method to retain previous knowledge in a network is elastic weight consolidation (EWC) \cite{Kirkpatrick2017}. EWC is a regularization technique aiming to constrain parameters of the model which are critical for performing previous tasks during the training of new tasks. Alternative methods attempt to overcome catastrophic forgetting by rehearsing past examples \cite{Lopez-Paz2017} or proxy information (pseudorehearsal) \cite{Ravishankar2019} when new information is added \cite{Robins1995CatastrophicPseudorehearsal}. 
%
In the field of medical imaging, continuous learning has been demonstrated to reduce catastrophic forgetting on segmentation and classification tasks. Karani et al. proposed domain-specific batch norm layers to adapt to new domains (different MR protocols) while learning segmentations of various brain regions \cite{Karani2018}. Baweja et al. applied EWC to sequentially learn normal brain structure and white matter lesion segmentation \cite{Baweja2018}.
Ravishankar et al. propose a pseudorehearsal technique and training of task-specific dense layers for pneumothorax classification \cite{Ravishankar2019}. 
%
These current approaches expect that the information about the domain to which a training example belongs, is available to the learning system. In real world medical imaging data, such information may not be available at the image level (e.g. a change in treatment policies, or different hardware updates across departments). At the same time, changes in protocol or scanner manufacturer may not automatically lead to a loss of performance of the model and considering each protocol as a novel domain may lead do adverse effects such as overfitting.

\paragraph{Contribution}
We propose an approach for continuous learning of continuously or repeatedly shifting domains. This is in contrast to most previous methods treating updates as sequentially adding well specified tasks. In contrast to existing rehearsal methods, we propose a technique that automatically infers data shifts without explicit knowledge about them. To this end, the method maintains a diverse memory of previously seen examples that is dynamically updated based on high level representations in the network. 

\section{Method}
We continuously update the parameters of an already trained model with new training data. Our approach composes this training data to capture novel data characteristics while sustaining the diversity of the overall training corpus. It chooses examples from previously seen data (\textit{dynamic memory $\mathcal{M}$}) and new examples (\textit{input-mini-batch $\mathcal{B}$}) to form the training data (\textit{training-mini-batch $\mathcal{T}$}) for the model update.

Our approach is a \textit{rehearsal method} to counter catastrophic forgetting in continuous learning. We adopt a \textit{dynamic memory (DM)} 
\begin{equation}
\mathcal{M} = \{\langle\mathbf{m}_1,{n}_1\rangle, \dots, \langle\mathbf{m}_M,n_M\rangle\} 
\end{equation}
 of a fixed-size $M$ holding image-label pairs $\langle\mathbf{m},{n}\rangle$ that are stored and eventually replaced during continuous training. To alleviate catastrophic forgetting, a subset of cases of $\mathcal{M}$ is used for rehearsal during every update step. It is critical that the diversity of $\mathcal{M}$ is representative of the visual variation across all tasks, even without explicit knowledge about the task membership of training examples. As the size of $\mathcal{M}$ is fixed, the most critical step of such an approach is to decide which samples to keep in $\mathcal{M}$ and which to replace with a new sample. To this end, we define a memory update strategy based on following rules: (1) every novel case will be stored in the memory, (2) a novel case can only replace a case in memory of the same class and (3) the case in memory that will be replaced is close according to a high level metric. Rule 1 allows the memory to dynamically adapt to changing variation. Rule 2 prevents class imbalance in the memory and rule 3 prevents the replacement of previous cases if they are visually distant. The metric used in rule 3 is critical as it ensures that cases of previous tasks are kept in memory and not fully replaced over time. 
We define a high-level metric based on the gram matrix $G^l \in \mathbb{R}^{N_l \times N_l}$ where $N_l$ is the number of feature maps in layer $l$.
$G_{ij}^l(\mathbf{x})$ is defined as the inner product between the vectorized activations $\mathbf{f}_{il}(\mathbf{x})$ and $\mathbf{f}_{jl}(\mathbf{x})$ of two feature maps $i$ and $j$ in a layer $l$ given a sample image $\mathbf{x}$:
\begin{equation}
\displaystyle G_{ij}^l(\mathbf{x}) = \frac{1}{N_lM_l} \mathbf{f}_{il}(\mathbf{x})^\top\mathbf{f}_{jl}(\mathbf{x})
\label{eq:grammatrix}
\end{equation}
where $M_l$ denotes the number of elements in the vectorized feature map (width $\times$ height).
For a set of convolution layers $\mathcal{L}$ we define a gram distance $\delta(\mathbf{x},\mathbf{y})$ between two images $\mathbf{x}$ and $\mathbf{y}$ as:
\begin{equation}
    \delta(\mathbf{x},\mathbf{y}) = \sum_{l \in \mathcal{L}} \frac{1}{N_l^2} \sum_{i = 1}^{N_l}\sum_{j = 1}^{N_l} (G_{ij}^l(\mathbf{x}) - G_{ij}^l(\mathbf{y}))^2
\end{equation}

The rationale behind using the gram matrix is the fact, that the gram matrix encodes high level style information. Here, we are interested in this style information to maintain a diverse memory not only with respect to the content but also with respect to different visual appearances. Similar gram distances have been used in computer vision methods in the area of neural style transfer as a way to compare the style of two natural images \cite{Gatys2016}. 

During continuous training, a memory update is performed after an input-mini-batch $\mathcal{B} = \{\langle\mathbf{b}_1,{c}_1\rangle, \dots, \langle\mathbf{b}_B,c_B\rangle\}$ of $B$ sequential cases (image $\mathbf{b}$ and label $c$) is taken from the data stream. Sequentially, each element of $\mathcal{B}$ replaces an element of $\mathcal{M}$. More formally, given an input sample $\langle \mathbf{b}_i, c_i\rangle$, the sample will replace the element in $\mathcal{M}$ with index 
\begin{equation}
 \xi(i) = \underset{j}{\text{arg\,min}}\,\delta(\mathbf{b}_i,\mathbf{m}_j)\ |\ c_i = n_j,\ j \in \{1,\dots,M\}.
\end{equation}
During the initial phase of continuous training, the memory is filled with elements of the data stream. Only after the desired proportion of a class in the memory is reached, the replacement strategy is applied. After the memory is updated, a model update is done by assembling a training-mini-batch $\mathcal{T} = \{\langle\mathbf{t}_1,{u}_1\rangle, \dots, \langle\mathbf{t}_T,u_T\rangle\}$ of size $T$. Each misclassified element of $\mathcal{B}$ for which the model predicted the wrong label is added to $\mathcal{T}$ and additional cases are randomly drawn from $\mathcal{M}$ until $|\mathcal{T}| = T$. Finally, using the training-mini-batch $\mathcal{T}$, a forward and backward pass is performed to update the parameters of the model.

\section{Experiments and Results}
We evaluated and studied the DM method in a realistic setting using medical images from clinical routine. To this end, we collected a representative dataset described in Section \ref{sec:dataset}. Based on these data, we designed a learning scenario with three tasks (A, B and C) in which a classifier pre-trained on task A is continuously updated with changing input data over time. Figure \ref{fig:exp_setup} illustrates the learning scenario, the data, and the experimental setup. Within the continuous dataset, we created two shifts, (1) a \textit{modality-shift} between scanner protocols and (2) a \textit{target-shift} by changing the target structure from high to low intensity. 

\begin{wraptable}{r}{7.5cm}
    \vspace{-2em}
    \setlength\intextsep{0pt}
    \centering
    \begin{tabular}{l|l|l|l|l}
        \centering
        {} & Task A & Task B & Task C & Total \\
        \hline 
Protocol         &   B3/3 &   B6/1 &   B6/1 &       \\
Target &    low &    how &   high &       \\
\hline \hline
Base             &   1513 &      0 &      0 &  1513 \\
Continuous       &   1513 &   1000 &   2398 &  4911 \\
Validation       &    377 &    424 &    424 &  1225 \\
Test             &    381 &    427 &    426 &  1234 \\
        \hline
    \end{tabular}
    \caption{\textbf{Data:} Splitting of the data into a base, continuous, validation, and test set. The number of cases in each split are shown.}
    \label{tbl:data}
    \vspace{-2em}
\end{wraptable}

\subsection{Dataset}
\label{sec:dataset}
In total, we collected 8883 chest CT scans from individual studies and for each extracted an axial slice at the center of the segmented \cite{Hofmanninger2020lungseg} lung. Each scan was performed on a Siemens scanner with either B3 reconstruction kernel and 3mm slice-thickness (B3/3) or B6 reconstruction kernel and 1mm slice-thickness (B6/1). We collected 3784 cases with B3/3 protocol and 5099 cases with the B6/1 protocol. We imprinted a synthetic target structure in the form of a cat on random locations, rotations and varying scale at 50\% of the cases (see also Figure \ref{fig:exp_setup}). The high-intensity target structures were engraved by randomly adding an offset between 200 and 400 hounsfield units (HU) and the low-intensity target structures by subtracting between 200 and 400 HU. A synthetic target was chosen to facilitate data set collection, as well as to create a dataset without label noise. Table \ref{tbl:data} lists the data collected and shows the partitioning into base, continuous, validation and test split and the stratification into the three tasks. 

\subsection{Experiments}
\label{sec:experiments}
We created the base model by fine-tuning  a pre-trained Res-Net50 \cite{He2015DeepRecognition} model, as provided by the pytorch-torchvision library\footnote{https://pytorch.org} on the \textit{base} training set (Task A).
Given this base model, we continuously updated the model parameters on the \textit{continuous} training set using different strategies:

\begin{itemize}
    \item \textbf{Naive}: 
    The naive approach serves as a baseline method by training sequentially on the data stream, without any specific strategy to counter catastrophic forgetting.
    \item \textbf{Elastic Weight Consolidation (EWC)}: As a reference method, we used EWC as presented in \cite{Kirkpatrick2017}. EWC regularizes weights that are critical for previous tasks based on the fisher information. We calculated the fisher information after training on the base set (Task A) so that in further updates, weights that are important for this task are regularized.
    \item \textbf{EWC-fBN}: In preliminary experiments, we found, that EWC is not suitable for networks using batch norm layers in scenarios where a modality-shift occurs. The reason for that is, that the regularization of EWC does not effect batch norm parameters (mean and variance) and that these parameters are constantly updated by novel data. Thus, to quantify this effect, we show results where we fixed the batch norm layers once the base training was completed.
    \item \textbf{Dynamic Memory (DM)}: The method as described in this paper. The input-mini-batch size $B$ and training-mini-batch size $T$ have been set to 8 for all experiments. If not stated differently, results have been computed with a memory size $M$ of 32. The gram matrices have been calculated on feature maps throughout the network covering multiple scales. Specifically, we used the feature maps of the last convolution layers before the number of channels is increased. That is, for Res-Net50 we calculate the matrices on four maps with 256, 512, 1024 and 2048 features.
\end{itemize}
In addition to the continuous scenario we trained an upper bound network in a conventional, epoch-based way, using all training data at once (\textit{full training}). All training processes were run five times to assess the variability and robustness and the results were averaged. All models were trained using an Adam optimizer \cite{Kingma2015Adam:Optimization} and binary cross entropy as a loss function for the classification task.
\\

\noindent With the described methods we,
\begin{itemize}
\vspace{-0.2cm}
    \item studied the \textbf{dynamics during training} by  calculating the classification accuracies on the validation set every 30 iterations.
    \item calculated \textbf{quantitative results} after training on the test set for each task separately. Classification accuracy, backward transfer (BWT) and forward transfer (FWT), as defined in \cite{Lopez-Paz2017} were used. BWT and FWT measure the influence that learning a new task has on the performance of previously learned, respectively future tasks. Thus, larger BWT and FWT values are preferable. Specifically, negative BWT quantifies catastrophic forgetting.
    \item studied the \textbf{influence of memory size} $M$ on the performance of our method during the training process and after training. We included experiments with $M$ set to 16, 32, 64, 80, 128 and 160.
\end{itemize}

\subsection{Results}

\begin{figure}[t!]
    \centering
    \includegraphics[width=0.8\textwidth]{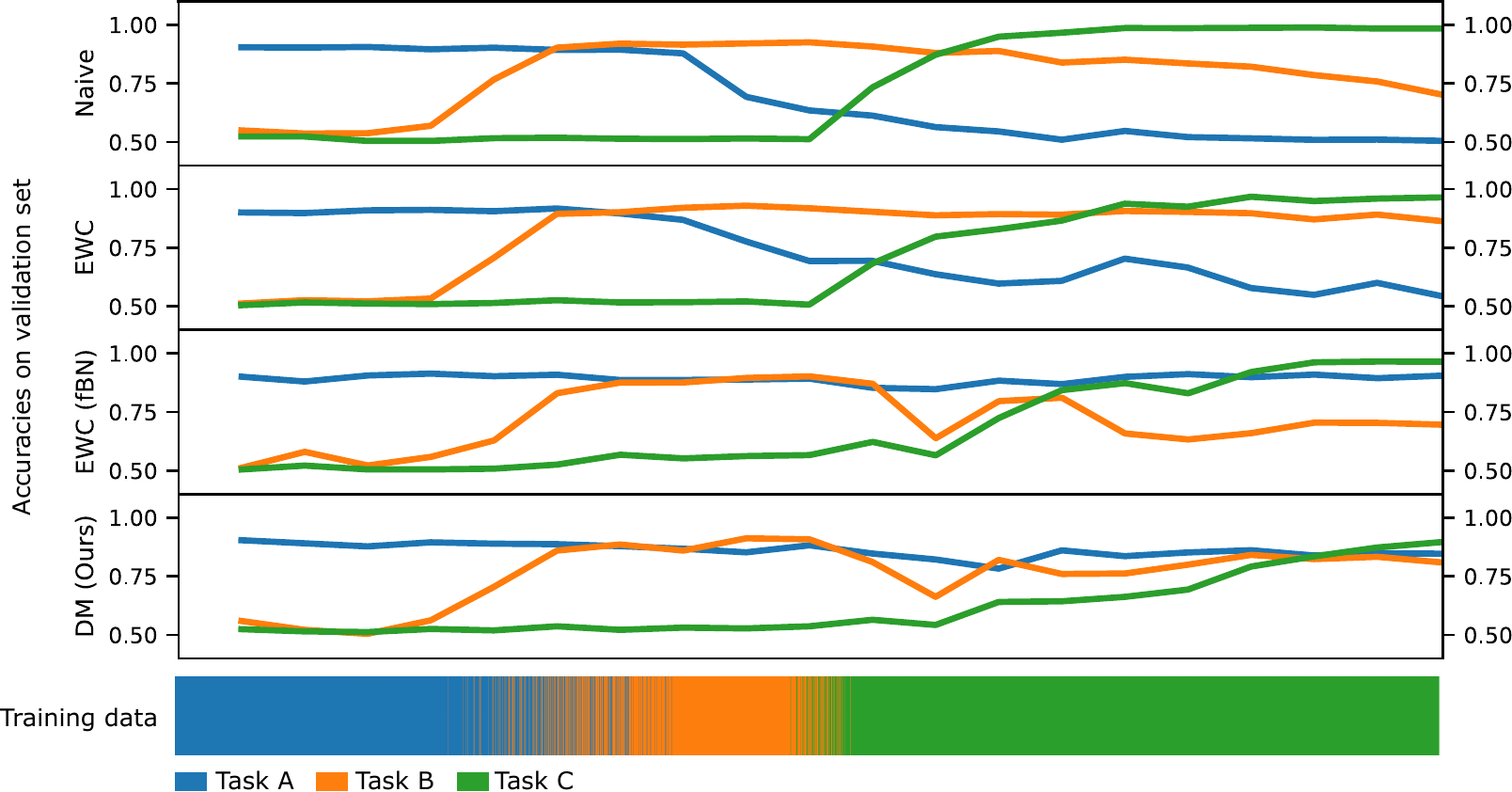}
    \caption{\textbf{Validation accuracy during training:} The curves show the change of validation accuracy for each of the tested approaches. Accuracies are computed on the validation sets for the three tasks during training. The last row show how the composition of the training data stream is changing over time.}
    \label{fig:res_lines}
    \vspace{-1em}
\end{figure}

\textbf{Dynamics during training} are shown in Figure \ref{fig:res_lines}. As expected, the naive approach shows catastrophic forgetting for task A (drop of accuracy from 0.91 to 0.51) and B (drop from 0.93 to 0.70) after new tasks are introduced. Without any method to counteract forgetting, knowledge of previous tasks is lost after sequential updates to the model. EWC exhibits catastrophic forgetting for task A (0.92 to 0.54) after the introduction of task B and C data into the data stream. The shift from task A to B is a modality shift, the batch norm layers of the network adapt to the second modality and knowledge about the first modality is lost. Although EWC protected the weights that are important for task A, those weights were not useful after the batch norm layers were adapted.
EWC-fBN avoids this problem by fixing the batch norm layers together with the weights that are relevant for task A. In this setting a forgetting effect for task B (accuracy drop from 0.90 to 0.70) can be observed after the target-shift to task C. This effect is due to the requirement of EWC to know when shifts occur. Thus, in our scenario, EWC only regularizes weights that are important for task A. As described previously, DM does not have this requirement by dynamically adapting to changing data. Using the DM approach only mild forgetting occurs and all three tasks reach a comparable performance of about 0.85 accuracy after training.

\begin{table}[b]
\begin{center}
\begin{tabular}{ l|r|r|r|r|r }
  & ACC Task A & ACC Task B & ACC Task C & BWT & FWT\\ \hline \hline
  Naive & $\otimes$ $0.51\pm0.00$ & $\otimes$  $0.71\pm0.02$ & $0.98\pm0.01$ & $-0.32\pm0.01$ & $0.05\pm0.00$\\ \hline
  EWC &  $\otimes$ $0.57\pm0.00$ & $0.83\pm0.00$ &   $0.91\pm0.00$ & $-0.20\pm0.00$ & $0.04\pm0.00$ \\ \hline
  EWC-fBN & $0.89\pm0.02$ & $\otimes$  $0.74\pm0.05$ &  $0.94\pm0.04$ & $-0.08\pm0.02$ & $0.06\pm0.00$\\ \hline
  DM (Ours) & $0.81\pm0.02$ &  $0.85\pm0.02$ & $0.92\pm0.04$ & $-0.07\pm0.02$ & $0.06\pm0.01$\\ \hline
  Full training & $0.92\pm0.01$ & $0.91\pm0.02$ & $0.97\pm0.00$ & - & -\\ \hline
\end{tabular}
\caption{Accuracy and BWT/FWT values for our dynamic memory (DM) method compared to baseline methods. Results were calculated on the test set after continuous training. Lower values marked with $\otimes$ indicate forgetting.}\label{tab:results}
\end{center}
\end{table}
\noindent \textbf{Quantitative results} are shown in Table \ref{tab:results}. The large negative BWT values for the naive approach (-0.32) and EWC (-0.20) indicate that these methods suffer from catastrophic forgetting. Using EWC-fBN mitigates the forgetting for task A, but the model is forgetting part of the knowledge for task B when task C is introduced (observable in Figure \ref{fig:res_lines}). Both, DM and EWC-fBN show comparable backward and forward transfer capabilities.
The accuracy values in Table \ref{tab:results} show that DM performs equally well on all tasks, while the other approaches show signs of forgetting of task A (naive and EWC) and task B (naive and EWC-fBN).
\begin{figure}[t]%
  \centering
  \subfloat[][]{\includegraphics[width=0.56\textwidth]{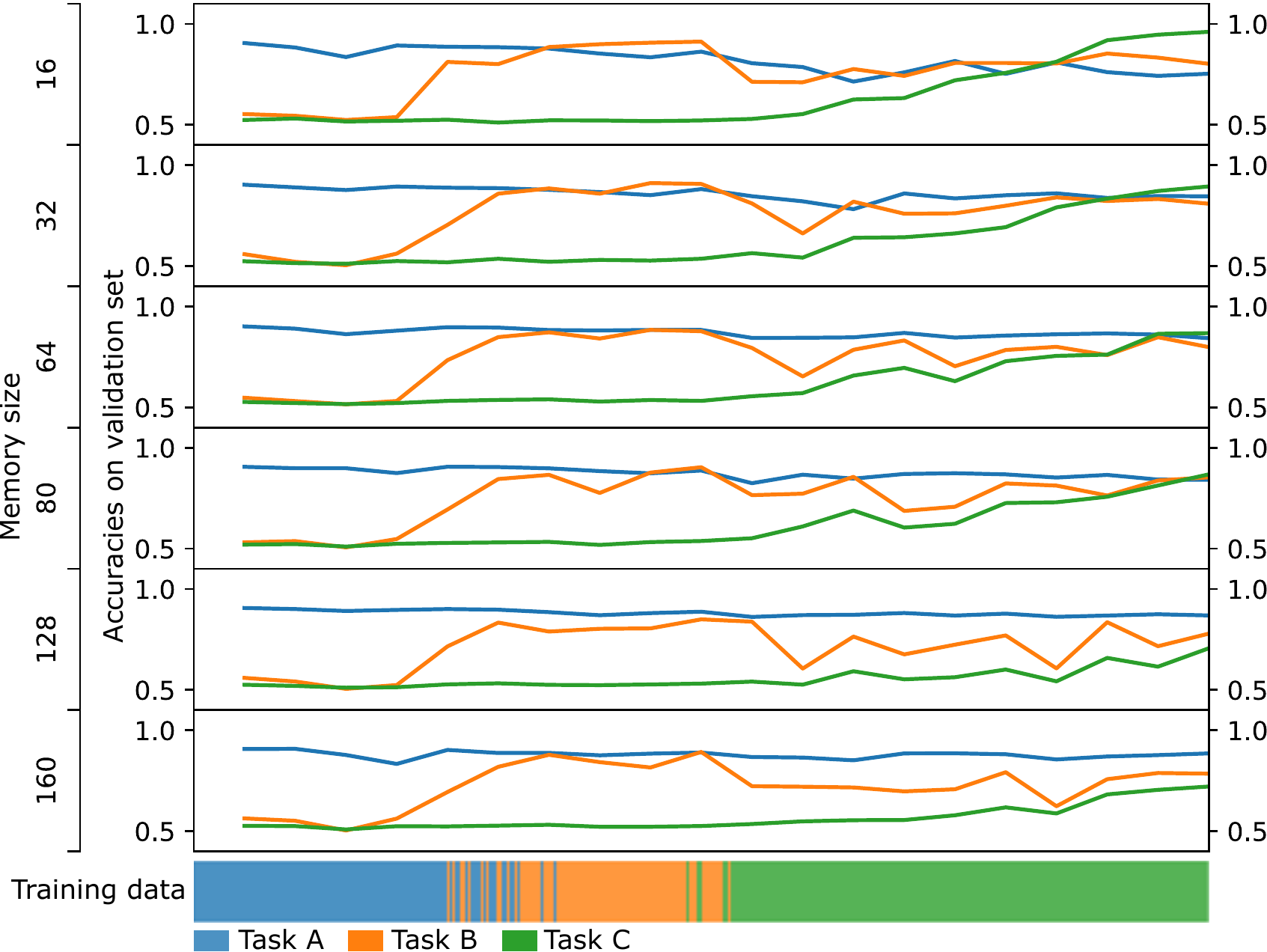}
  \label{fig:memorysizea}}%
  \qquad
  \subfloat[][]{\small
  \begin{tabular}[b]{l|r|r|r|r}
    $M$ &  Task A &  Task B &  Task C &  Avg \\
    \hline
    16 &    0.73 &    0.83 &    0.96 &      0.84 \\ \hline
    32 &    0.82 &    0.86 &    0.93 &      0.87 \\ \hline
    64 &    0.85 &    0.81 &    0.87 &      0.84 \\ \hline
    80 &    0.86 &    0.82 &    0.87 &      0.85 \\ \hline
    128 &    0.87 &    0.75 &    0.71 &      0.78 \\ \hline
    160 &    0.88 &    0.79 &    0.69 &      0.79 \\
    \hline
    \end{tabular}\label{fig:memorysizeb}}%
  \caption{\textbf{Memory size: (a)} The effect of memory size during training with DM. Small memory sizes tend to catastrophic forgetting, while large memory sizes lead to slow training of later tasks. \textbf{(b)} Classification accuracy on the validation set after training for varying memory sizes.}
  \label{fig:memorysizeab}
  \vspace{-1em}
\end{figure}
 
\noindent \textbf{The influence of memory size} during training is shown in Figure \ref{fig:memorysizea}. For a small $M$ of 16, adapting to new tasks is fast. However, such a limited memory can only store a limited data variability leading to catastrophic forgetting effects. Increasing memory size decreases catastrophic forgetting effects but slows adaption of new tasks. For the two largest investigated sizes 128 and 160  adaption, especially after the target-shift (Task C), is slower. The reason is, that more elements of task A are stored in the memory and more iterations are needed to fill the memory with samples of task C. This reduces the probability that elements from task C are drawn and in turn, slows down training for task C. For the same reason, with higher memory sizes there is an increase in task A accuracy, since examples from the first task are more often seen by the network. Results indicate that setting the memory size is a trade-off between faster adaption to novel data and more catastrophic forgetting.
In our setting, training worked comparably well on $M$ of 32, 64, and 80. Considering memory sizes between 16 and 160, setting $M=32$ resulted in the highest average accuracy of 0.87 on the validation set (Fig. \ref{fig:memorysizeb}). Therefore, we used this model for comparison to other continuous learning methods on the test set.
\section{Conclusion}
Here, we presented a continuous learning approach to deal with modality- and tasks-shifts induced by changes in protocols, parameter setting or different scanners in a clinical setting. We showed that maintaining a memory of diverse training samples mitigates catastrophic forgetting of a deep learning classification model. We proposed a memory update strategy that is able to automatically handle shifts in the data distribution without explicit information about domain membership and the moment such a shift occurs.

\subsubsection*{Acknowledgments}
This work was supported by Austrian Science Fund (FWF) I 2714B31 and Novartis Pharmaceuticals Corporation.
%
%
%
\bibliographystyle{splncs04}
\bibliography{continuouslearning}
\end{document}